\documentclass[english,a4paper,11pt]{article}
\usepackage{fullpage}

\usepackage{amsxtra, amsfonts, amssymb, amstext}
\usepackage{amsthm}
\usepackage{booktabs}
\usepackage{longtable}
\usepackage{nicefrac}
\usepackage{xspace}
\usepackage[noadjust]{cite}
\usepackage{url}
\usepackage{graphics,color}
\usepackage[colorlinks]{hyperref}
\definecolor{linkblue}{rgb}{0.1,0.1,0.8}
\hypersetup{colorlinks=true,linkcolor=linkblue,filecolor=linkblue,urlcolor=linkblue,citecolor=linkblue}
\usepackage[algo2e,ruled,vlined,linesnumbered]{algorithm2e}
\newcommand{\assign}{\leftarrow}
\usepackage{graphicx}
\usepackage{wrapfig}
\usepackage{balance}
\usepackage{caption}
\usepackage{subcaption}

\usepackage{rotating}
\usepackage{arydshln}

\newcommand{\ignore}[1]{}

\allowdisplaybreaks[1]

\newcommand{\R}{\mathbb{R}}

\renewcommand{\epsilon}{\varepsilon}

\newcommand{\E}{\mathbb{E}}
\renewcommand{\Pr}{\mathbb{P}}

\DeclareMathOperator{\flip}{flip}
\DeclareMathOperator{\opt}{opt}
\DeclareMathOperator{\drift}{drift}

\DeclareMathOperator{\Bin}{Bin}

\newcommand{\onemax}{\textsc{OneMax}\xspace}

\newcommand{\OM}{\textsc{Om}\xspace}

\newcommand{\oea}{$(1 + 1)$~EA\xspace}
\newcommand{\oeaopt}{$(1 + 1)$~EA$_{\text{opt}}$\xspace}
\newcommand{\oeadrift}{$(1 + 1)$~EA$_{\text{drift}}$\xspace}

\newcommand{\oeares}{$(1 + 1)$~EA$_{>0}$\xspace}
\newcommand{\oearesopt}{$(1 + 1)$~EA$_{>0,\text{opt}}$\xspace}
\newcommand{\oearesdrift}{$(1 + 1)$~EA$_{>0,\text{drift}}$\xspace}

\newcommand{\RLSopt}{RLS$_{\text{opt}}$\xspace}
\newcommand{\RLSdrift}{RLS$_{\text{drift}}$\xspace}

\DeclareMathOperator{\RLS}{RLS}
\DeclareMathOperator{\EA}{EA}
\DeclareMathOperator{\fRLSopt}{\RLS_{\text{opt}}\xspace}
\DeclareMathOperator{\fRLSdrift}{\RLS_{\text{drift}}\xspace}

\DeclareMathOperator{\foearesopt}{(1 + 1)~\EA_{>0,\opt}\xspace}
\DeclareMathOperator{\foearesdrift}{(1 + 1)~\EA_{>0,\drift}\xspace}

\DeclareMathOperator{\foeaopt}{(1 + 1)~\EA_{\opt}\xspace}
\DeclareMathOperator{\foeadrift}{(1 + 1)~\EA_{\drift}\xspace}

\newcommand{\kopt}{k_{\text{opt}}}
\newcommand{\kdrift}{k_{\text{drift}}}
\newcommand{\popt}{p_{\text{opt}}}
\newcommand{\pdrift}{p_{\text{drift}}}
\newcommand{\presopt}{p_{>0,\text{opt}}}
\newcommand{\presdrift}{p_{>0,\text{drift}}}

\newcommand{\pmin}{p_{\min}}

\title{Maximizing Drift is Not Optimal for Solving OneMax}

\author{Nathan Buskulic$^1$ and Carola Doerr$^1$}
\date{
$^1$Sorbonne Universit\'e, CNRS, Laboratoire d'Informatique de Paris 6, Paris, France\\ 
\vspace{1.5ex}
\today
}

\begin{document}
\maketitle

\begin{abstract}
It seems very intuitive that for the maximization of the OneMax problem $\OM(x):=\sum_{i=1}^n{x_i}$ the best that an elitist unary unbiased search algorithm can do is to store a best so far solution, and to modify it with the operator that yields the best possible expected progress in function value. This assumption has been implicitly used in several empirical works. In [Doerr, Doerr, Yang: \emph{Optimal parameter choices via precise black-box analysis}, TCS, 2020] it was formally proven that this approach is indeed almost optimal. 

In this work we prove that drift maximization is not optimal. More precisely, we show that for most fitness levels between $n/2$ and $2n/3$ the optimal mutation strengths are larger than the drift-maximizing ones. This implies that the optimal RLS is more risk-affine than the variant maximizing the step-wise expected progress. We show similar results for the mutation rates of the classic {(1+1)} Evolutionary Algorithm (EA) and its resampling variant, the {(1+1)} EA$_{>0}$.  

As a result of independent interest we show that the optimal mutation strengths, unlike the drift-maximizing ones, can be even.
\end{abstract}

\sloppy{
\section{Introduction}
\label{sec:introduction}

It is well understood that iterative optimization heuristics like local search variants, evolutionary algorithms, estimation of distribution algorithms, etc.~can benefit from non-static choices of the parameters that determine their search radius, population size, or selective pressure. The question how to select these parameters dynamically is the subject of \emph{parameter control,} which studies different techniques to achieve a good fit between suggested and optimal parameter values. 

Complementing a diverse body of empirical works demonstrating advantages of parameter control mechanisms~\cite{KarafotiasHE15,AletiM16}, there is an increasing interest in proving such benefits by mathematical means~\cite{DoerrD18chapter}. Among the significant advances in this direction are, in chronological order (with respect to the conference announcements), the analysis of a success-based adaptation strategy for the choice of the offspring population size $\lambda$ of the $(1+\lambda)$~EA in distributed models of computation~\cite{LassigS11}, the self-adjusting $(1+(\lambda,\lambda))$~Genetic Algorithm (GA) using the one-fifth success rule~\cite{DoerrD18ga}, a learning-based selection of the search radii in Randomized Local Search~\cite{DoerrDY16PPSN}, 
and the self-adjusting~\cite{DoerrGWY19} and self-adaptive~\cite{DoerrWY18} mutation rates in a $(1+\lambda)$ and $(1,\lambda)$~Evolutionary Algorithm (EA), respectively. All these references consider the optimization of \onemax, the problem of maximizing the counting-ones function $\OM:\{0,1\}^n \to \R, x \mapsto \sum_{i=1}^n{x_i}$. 
Only few theoretical results analyzing algorithms with adaptive parameters consider different functions, e.g.,~\cite{LissovoiOW20,DoerrLOW18,DoerrDK18} (see~\cite{DoerrD18chapter} for a complete list of references). \onemax also plays a prominent role in empirical research on parameter control. In both communities, it is argued that the consideration of \onemax provides a ``sterile EC-like environment''~\cite{FialhoCSS08}, in which the optimal parameter values are well understood. 

In light of the existing literature it is interesting to note that most works, implicitly or explicitly, assume that for the considered algorithms the optimal strategy for the maximization of \onemax is a greedy selection of the best so far solution, and the variation of the same by the mutation rate/step size that maximizes the expected gain in function value~\cite{Back92,Back93,FialhoCSS08,FialhoCSS09}. Thierens~\cite{Thierens09} explicitly argues that a particularly useful property of \onemax, which makes this problem a very suitable benchmark for adaptive operator selection, is the fact that the reward of an operator can be computed exactly. He then proceeds by comparing the step-wise expected fitness gains made by different operators, and ranks operators by this value. He thus uses as underlying assumption that drift-maximization is optimal. 

That this widely believed-to-be-optimal \emph{drift-maximizing} strategy is indeed almost optimal was formally proven in~\cite{DoerrDY20}. More precisely, it is shown in~\cite{DoerrDY20} that the best unary unbiased black-box algorithm for \onemax cannot be better by more than an additive $o(n)$ term than the RLS variant that flips in each iteration the drift-maximizing number of bits in a best-so-far solution. Both algorithms have an expected optimization time $n \ln(n) - cn \pm o(n)$, for a constant $c$ between $0.2539$ and $0.2665$. 

It was conjectured in~\cite[Section~3.1]{DoerrW18} that the drift-maximizing RLS is not only ``almost'' optimal, but indeed optimal. As mentioned, this conjecture was also---explicitly or implicitly---made in the empirical works cited above (and several other works on the \onemax function). We show in this work that this conjecture is false. More precisely, we show that maximizing drift is not optimal neither for RLS nor for the {(1+1)}~EA nor for its resampling variant, the \oeares, suggested in~\cite{CarvalhoD17}. 

We explain where the difference between optimal and drift-maximizing strategies comes from, define precisely how to obtain the optimal mutation rates, numerically compute these for some selected dimensions up to $n=10{,}000$, and analyze the differences between drift-maximizing and optimal mutation rates. We also compare the performances of optimal and drift-maximizing algorithms, and show that the differences in mutation rates/step sizes---albeit significant---result only in marginal differences in terms of overall running time. Given the above-mentioned results in~\cite{DoerrDY20}, the last statement is not surprising. The main contribution of our work is therefore not to be found in tremendous performance gains, but in new structural insights for the optimization of \onemax, the arguably most widely used benchmark for parameter control and adaptive operator selection mechanisms. 

We note that the argument \emph{why} drift-maximization is not optimal is quite easy to understand. Basically, our result is built upon the observation that the drift-maximizer values a potential fitness progress of $i$ by exactly this gain. More precisely, in the computation of the drift, the probability of creating an offspring $y$ of $x$ is multiplied by the difference $\max\{0,\OM(y)-\OM(x)\}$, for each possible offspring $y$. The optimal algorithms, however, value a fitness gain of $i$ by the gain in the expected remaining running time. Since this difference in expected remaining running time is much larger than the fitness difference, the optimal RLS and \oea variants use mutation rates that are larger than the drift-maximizing ones. Put differently, they trade a smaller expected progress for a slightly larger probability of making a larger fitness gain. That is, the optimal algorithms are more risk-affine than the drift-maximizing ones. This quite intuitive fact seems to have been overlooked in the evolutionary computation (EC) community.

Our work has recently been extended to $(1+\lambda)$-type RLS and EAs~\cite{BuzdalovD20}. In that work, not only the optimal mutation rates are computed, but also the expected remaining running times for sub-optimal mutation rates -- information that can be used to identify weak spots of parameter control mechanisms.  

\textbf{Precise Running Time Bounds.} While we focus in this work on very precise running time bounds for concrete problem dimensions, which we compute numerically, we note that there exists a significant body of related theoretical works, which focus on asymptotically optimal mutation rates and running times. In addition to the works mentioned above, which all deal with adaptive parameter schemes, we consider the following ones particularly interesting in the context of our study. For the classic RLS variant, which always flips exactly one bit in each iteration, the expected running time on \onemax was computed very precisely in~\cite{DoerrD16impact}. For the \oea with static mutation rate $1/n$, the best known bounds are proven in~\cite{HwangPRTC18} and in the recent work~\cite{HwangW19}, which are precise up to an additive $O(\log (n)/n)$ and $O(\log n)$ term, respectively. For other static mutation rates, the best known results are available in~\cite{Witt13j}. 

\textbf{Online Repository.} Codes and details for the here-described algorithms can be found on the GitHub page of this project at \url{https://github.com/NathanBuskulic/OneMaxOptimal}. 
The interested reader can find there not only the performance data, but also the drift-maximizing and optimal step sizes/mutation rates of the algorithms discussed below, for problem dimensions up to $n=10{,}000$.

\section{The OneMax Problem}
\label{sec:OM}

\onemax, also referred to as counting-ones problem in the early works on evolutionary computation, is the problem of maximizing the function $$\OM:\{0,1\}^n \to \R, x \mapsto \sum_{i=1}^n{x_i},$$ which simply assigns to each bit string the number of ones in it. \onemax is considered to be one of the ``easiest'' non-trivial benchmark problems, for two reasons. Firstly, a number of results exist that show that for several (classes of) algorithms the expected optimization time on \onemax is not bigger than that on any other unimodal function of the same dimension, cf.~\cite{DoerrJW12,Sudholt13,CorusHJOSZ17} for examples. A second reason to declare \onemax as ``easy'', yet useful, benchmark problem is its (presumably) simple structure, which allows us to understand well the optimization process of classical optimization heuristics. One structural property that is particularly useful in runtime analyses is the perfect fitness-distance correlation; i.e., whenever $\OM(x)>\OM(y)$ for two search points $x$ and $y$, then the distance of $x$ to the optimum is strictly smaller than that of $y$. 

For readers wondering about the usefulness of a single benchmark instance, we note that for most evolutionary algorithms (EAs) and local search variants such as Randomized Local Search (RLS), Simulated Annealing, etc. the \onemax problem is identical to the problem of maximizing any of the functions $\OM_z:\{0,1\}^n \to \R, x \mapsto H(z,x):=|\{ i \in [n] \mid x_i \neq z_i\}|$, since for any $z \in \{0,1\}^n$ the Hamming distance problem $\OM_z$ has a fitness landscape that is isomorphic to that of \onemax, and the mentioned algorithms are oblivious of the exact problem representation. That is, \onemax is essentially just one representative of the class of Hamming distance problems.

\onemax is often termed the ``drosophila of EC'', because of the vast amount of literature studying this problem, both in empirical and in theoretical works. In the context of our study in particular the works~\cite{Back92,Back93,FialhoCSS08,FialhoCSS09,Thierens09,BadkobehLS14,DoerrD18ga,DoerrDY20,DoerrDY16PPSN,DoerrGWY19,DoerrWY18,LaillevaultDD15,DoerrW18} are worth mentioning, as they all study the benefits of non-static parameter choices on this problem, for different local search variants and evolutionary algorithms. Among these works, the empirical ones focus on operators that maximize the expected progress (``drift'') per each round, either without further justifying it, or explicitly mentioning that drift-maximization is optimal (an assumption that we will refute in Section~\ref{sec:example}). Among the theoretical works, most are interested in deriving asymptotic results only, with the only exception of~\cite{DoerrDY20,DoerrDY16PPSN}, where very precise bounds for the optimization time of two adaptive RLS variants are proven. Most relevant to our work is the mentioned result from~\cite{DoerrDY20} which proves that the drift-maximizing strategy mentioned above is indeed almost optimal. When we show in the next sections that the best possible RLS variant is not the drift-maximizing one, we know by the result from~\cite{DoerrDY20} that the gain in expected optimization time cannot be more than an additive $O(n^{2/3}\log^9 n)$ term. 

\section{Elitist {(1+1)} Unbiased Algorithms}
\label{sec:algos}

 \begin{algorithm2e}[t]%
	\textbf{Initialization:} 
	Sample $x \in \{0,1\}^{n}$ uniformly at random and compute $f(x)$\;
  \textbf{Optimization:}
	\For{$t=1,2,3,\ldots$}{
		\label{line:ellres}Sample $k \sim D(n,f(x))$\;
		$y \assign \flip_k(x)$\;
		evaluate $f(y)$\;
		\lIf{$f(y)\geq f(x)$}{$x \assign y$}	
}
\caption{Blueprint for elitist {(1+1)} unbiased black-box algorithms maximizing a function $f:\{0,1\}^n \to \R$}
\label{alg:11}
\end{algorithm2e}

We are concerned in this work with algorithms following the blueprint given in Algorithm~\ref{alg:11}. These algorithms start the optimization in a randomly chosen solution $x$. In each iteration exactly one offspring $y$ is sampled by first copying $x$ and then flipping the entries of $k$ randomly chosen, pairwise different positions $i_1, i_2, \ldots, i_{k}$. The \emph{parent} $x$ is replaced by its \emph{offspring} $y$ if and only if $f(y) \ge f(x)$, i.e., if and only if the offspring is at least as good as $y$. Algorithms adhering to this scheme are referred to in the theory of EA literature as \emph{elitist unary unbiased black-box algorithms}~\cite{DoerrL17ECJ}. 

Elitist unary unbiased black-box algorithms differ only in the choice of the \emph{mutation strength}~$k$. The two most commonly studied classes of algorithms are \textbf{Randomized Local Search (RLS)} variants, which use a \emph{deterministic} choice of $k$, and \textbf{{(1+1)} Evolutionary Algorithms (EAs)}, which sample $k$ from $\Bin(n,p)$, i.e., from a binomial distribution with $n$ trials and success rate $p$. We note that traditionally \emph{constant} choices, $k=1$ for RLS and $p=1/n$ for the \oea, are studied, but here in this work we focus on \emph{non-static} mutation strengths $k$ and \emph{mutation rates} $0\le p \le 1$. More precisely, we study fitness-dependent choices $k(\ell)$ and $p(\ell)$, which take into account the function value (\emph{fitness}) $\ell=\OM(x)$ of the current-best solution. In the terminology proposed in~\cite{DoerrD18chapter} such parameter control schemes classify as \emph{state-dependent,} since the parameter value depends only on the current-best solution but not on any other information about the optimization process. 
The objective of our work is to identify the functions $\ell \mapsto k(\ell)$ and $\ell \mapsto p(\ell)$ that minimize the expected running time of RLS and the \oea, respectively, when optimizing \onemax.  

We add to our investigation the \textbf{\oeares}, which samples $k$ from a conditional binomial distribution $\Bin_{>0}(n,p)$, which is defined by $\Bin_{>0}(n,p)(0)=0$ and $\Bin_{>0}(n,p)(i)=\Bin(n,p)(i)/(1-(1-p)^n) = \binom{n}{i}p^i(1-p)^{n-i}/(1-(1-p)^n)$ for $i \in [n]$. That is, the probability of the \oeares to flip $i$ bits equals that of the \oea conditional on flipping at least one bit. The \oeares was suggested in~\cite{CarvalhoD17} as an algorithm that more closely resembles common implementations of the \oea, cf. also discussions in~\cite{CarvalhoD18}. The \oeares can be seen as an intermediate algorithm between the RLS variant always flipping one bit and the (unconditional) \oea, since for $p$ converging to 0 the distribution $\Bin_{>0}(n,p)$ concentrates on 1, so that for small $p$ the behavior of the \oeares ``converges'' against that of RLS. 

We note that other elitist unary unbiased black-box algorithms have been recently introduced. The \emph{fast Genetic Algorithm (GA)} suggested in~\cite{FastGA17} samples the mutation strength~$k$ from a power-law distribution, and $k$ is sampled from a normal distribution $N(\mu,\sigma^2)$ in the \emph{normalized EA} studied in~\cite{YeDB19}. We will nevertheless focus in this work on RLS and {(1+1)} EA variants only, simply because they are still the most commonly studied algorithms in evolutionary computation. We note though that an extension of our work in particular to results covering the normalized EAs would be interesting, since this algorithm class can be seen as a meta-model between the class of RLS algorithms and the class of $(\mu+\lambda)$~EAs. 

\section{Maximizing Drift is Not Optimal}
\label{sec:example}

As mentioned in Section~\ref{sec:algos}, our main interest is in identifying the functions $\kopt:[0..n-1] \to [0..n]$, $\popt:[0..n-1] \to [0,1]$, and $\presopt: [0..n-1] \to [0,1]$ for which the following three algorithms have a best possible expected optimization time: 
 \begin{itemize}
	 \item \textbf{\RLSopt,} the RLS variant flipping in each iteration exactly $\kopt(\OM(x))$ bits (i.e., using the deterministic mutation strength~$\kopt(\OM(x))$),
	\item \textbf{\oeaopt,} the \oea variant using standard bit mutation with mutation rate $\popt(\OM(x))$ (i.e., the algorithm sampling the mutation strength from the binomial distribution $\Bin(n,\popt(\OM(x)))$), and 
	\item \textbf{\oearesopt,} the \oeares variant using conditional standard bit mutation flipping at least one bit with mutation rate $\presopt(\OM(x))$ (i.e., sampling the mutation strength from the conditional binomial distribution $\Bin_{>0}(n,\presopt(\OM(x)))$.
 \end{itemize}
Note that, formally, we should write $\kopt(n)$, $\popt(n)$, and $\presopt(n)$, since these functions depend on the dimension. However, we shall often omit the explicit mention of the dimensions in order to ease the notation. The same applies to the corresponding functions $\kdrift(n)$, $\pdrift(n)$, and $\presdrift(n)$.

It may be surprising that, after so many years of research on the \onemax problem, none of the three algorithms above has been explicitly computed. As mentioned in the introduction, there are two main reasons explaining this situation. Firstly, it is widely believed that the functions $\kdrift$, $\pdrift$, and $\presdrift$, which maximize in each step the expected fitness gain (\emph{drift}) of flipping $k =\kdrift(\OM(x))$, $k \sim \Bin(n,\pdrift(\OM(x)))$, and $k \sim \Bin_{>0}(n,\presdrift(\OM(x)))$ bits, respectively, are optimal. As already discussed, such claims can be quite frequently found in the literature~\cite{Back92,FialhoCSS08,DoerrW18}. We will show in this section that these claims are not correct, by presenting examples which demonstrate that better expected optimization times can be achieved by choosing $\kopt \neq \kdrift$, $\popt \neq \pdrift$, and $\presopt \neq \presdrift$, respectively. In Section~\ref{sec:opt} we will quantify the discrepancies between drift-maximizing and optimal (i.e., time-minimizing) functions for dimensions up to $n=10{,}000$. Section~\ref{sec:runtime} discusses the impact of these differences on the overall running time.   

\subsection{\texorpdfstring{$\fRLSopt \neq \fRLSdrift$}{RLSopt is not equal to RLSdrift} for \texorpdfstring{$n=3$}{n=3}}
\label{sec:RLS3}

We first show that $\kdrift \neq \kopt$. That is, we study the drift-maximizing and the time-minimizing variants of RLS, which we call \RLSdrift and \RLSopt in the following, and show that they are not identical. Interestingly, it suffices to regard $n=3$ for an example for which the two functions differ. The following table summarizes for $n=3$ the functions $\kdrift$, $\kopt$, and the expected remaining running times $\E[T_{\drift}(\ell)]$ and $\E[T_{\opt}(\ell)]$ for \RLSdrift and \RLSopt, respectively, when starting in a solution $x$ of fitness $\OM(x)=\ell$. In column $p^0(\ell)$ we list the probability that a random initial solution has fitness value $\ell$. Since uniform random initialization is used, $p^0(\ell)=\binom{n}{\ell}/2^n$. The last line provides the overall expected optimization time of both algorithms. Note that, by the law of total probability, 
$$\E[T]=1+\sum_{\ell=0}^n{p^0(\ell) \E[T(\ell)]},$$ where the ``+1''-term accounts for the evaluation of the initial solution.  

\vspace{2ex}
\begin{tabular}{l|l|cc|cc}
$\ell$                   & $p^0(\ell)$ & $\kdrift(\ell)$ & $\E[T_{\drift}(\ell)]$ & $\kopt(\ell)$ & $\E[T_{\opt}(\ell)]$ \\
\hline
3                        & $1/8$                      & -               & 0                        & -             & 0 \\
2                        & $3/8$                      & 1               & 3                     & 1             & 3  \\
1                        & $3/8$                      & 3           & 4                  & 2          & 3                   \\
0                        & $1/8$                      & 3               & 1                        & 3             & 1                      \\
\hline 
\multicolumn{2}{c|}{$\E[T]$} &   \multicolumn{2}{c|}{ 3.75}                    & \multicolumn{2}{c}{3.375}                 
\end{tabular} 
\vspace{2ex}

As we see from the last line, the overall expected running time of \RLSopt is 3.375 and thus strictly smaller than that of \RLSdrift, which is 3.75. We briefly explain how the entries in this table are computed. 

\textbf{Computation of \RLSdrift.} We start our explanation with the computation of $\kdrift(n):[0..n-1] \to [0..n]$ and $\E[T_{\drift}(\ell)]$. The function $\kdrift$ was defined above to be the one that maps each fitness value to the number of bits that need to be flipped in order to maximize the expected progress in fitness value, i.e., $\kdrift(n,\ell)$ is defined to be the value of $k$ that maximizes the expression
\begin{align}
\label{eq:Edriftk}
	&\E[\Delta(n,\ell,k)]  :=\\
	&\nonumber \E[ \max\{\OM(y)-\OM(x), 0\} \mid \OM(x)=\ell, y \assign \flip_{k}(x)] \\
= &\nonumber \sum_{i=\ell+1}^{\ell+k}{(i-\ell) \Pr[\OM(y)=i \mid \OM(x)=\ell, y \assign \flip_{k}(x)]} \\
= &\nonumber \sum_{i=\lceil k/2 \rceil}^{k}
	\frac{\binom{n-\ell}{i}\binom{\ell}{k-i}\left(2i-k\right)}{\binom{n}{k}},
\end{align}
where we use in the last line the fact that flipping $i$ of the $n-\ell$ previously incorrect bits implies that we flip $k-i$ of the $\ell$ previously correct bits, which results in a fitness increase of $i-(k-i)=2i-k$. This event occurs with probability $\frac{\binom{n-\ell}{i}\binom{\ell}{k-i}}{\binom{n}{k}}$, since there are $\binom{n-\ell}{i}$ different ways of choosing $i$ previously incorrect bits, $\binom{\ell}{k-i}$ ways of choosing $k-i$ previously correct bits, and $\binom{n}{k}$ ways of choosing $k$ pairwise different bit positions.  
When two or more values $k$ exist that minimize this expression, we follow the convention made in~\cite{DoerrDY20} and chose in all our computations below the smallest of these drift-maximizing mutation strengths, i.e., formally, 
$\kdrift(n,\ell)=\min \big\{\arg\max_k \E[ \Delta(\ell,k)] \big\}$.\footnote{In light of the results presented in this paper, it seems likely that for $\ell>n/2$ the better choice would be $\kdrift(n,\ell)=\max\{\arg\max_k \E[ \Delta(\ell,k)]\}$, but given the small discrepancies in the resulting running times (cf. Section~\ref{sec:runtime}) we do not investigate this question further.} 

It is easily seen that that for $n=3$ and $\ell=2$ flipping one bit is optimal, since this is the only mutation strength yielding positive drift. With this value of $\kdrift(n=3,\ell=2)$ the expected remaining time $\E[T_{\drift}(n=3,\ell=2)]$ to find the optimal solution is 3. 
For $\ell=1$, the expected progress of $\flip_1$, i.e., of flipping one bit, is $\Pr[\OM(y)=2 \mid \OM(x)=1, y \assign \flip_{1}(x)] = 2/3$, the expected progress of $\flip_2$ equals $2\Pr[\OM(y)=3 \mid \OM(x)=1, y \assign \flip_{2}(x)]=2/3$ (note here that no fitness gain of one is possible), and the expected progress of $\flip_3$ is $1$, since in this case we deterministically obtain an offspring $y$ of fitness $\OM(y)=2$. The best drift is thus obtained by operator $\flip_3$, which implies $\kdrift(n=3,\ell=1)=3$. With this choice of the mutation strength, the expected remaining optimization time equals $1+\E[T_{\drift}(n=3,\ell=2)]=4$. In general, $\E[T_{\drift}(n,\ell)]$ can be computed as 
\begin{align*}
1+\sum_{i=\ell+1}^{n-1}{\Pr[\OM(y)=i \mid \mathcal{E}] \E[T_{\drift}(n,i)] },
\end{align*}
where $\mathcal{E}$ is the event that $\OM(x)=\ell$ and $y \assign \flip_{\kdrift(n,\ell)}(x)$.
When $\OM(x)=0$ then flipping all bits, i.e., applying $\flip_n$ is optimal, since it directly produces the optimal solution. Note here that, more generally, the function value of the bitwise complement $\bar{x}$ of a solution $x$ equals $\OM(\bar{x})=n-\OM(x)$.

With these values, the expected running time of \RLSdrift on $n=3$ is equal to $1+\tfrac{3}{8}3+\frac{3}{8}4+\tfrac{1}{8}=\tfrac{15}{4}=3.75$.

\textbf{Computation of \RLSopt.} We next discuss how to compute $\kopt(n):[0..n-1] \to [0..n]$ and $\E[T_{\opt}(\ell)]$. This time, we start our investigation by recalling that, by the law of total probability, the expected optimization time $\E[T(\fRLSopt)]$ of \RLSopt equals 
$1+\sum_{\ell=0}^{n-1}{\Pr[\OM(x^0)=\ell] \E[T_{\opt}(\ell)]}$. The best-possible RLS algorithm is hence the one using at each fitness level $\ell$ the mutation strength~$\kopt(n,\ell)$ which minimizes the expected remaining optimization time $\E[T_{k}(n,\ell)]$ of flipping $k$ bits, which is equal to
\begin{align}\label{eq:kopt}
1+\sum_{i=\ell+1}^{n-1}{\Pr[\OM(y)=i \mid \OM(x)=\ell, y \assign \flip_{k}(x)] \E[T_{\opt}(n,i)]}.
\end{align}
Formally, we set again $k = \arg \min_k \E[T_{k}(n,\ell)]$. Note that the expression in~\eqref{eq:kopt} requires to know the values $\E[T_{\opt}(n,i)]$ for $i>\ell$. In order to compute $\kopt(n,\ell)$ one therefore has to start with fitness level $n-1$. Once $\kopt(n,n-1)$ and $\E[T_{\opt}(n,n-1)]$ are known, $\kopt(n,n-2)$ and $\E[T_{\opt}(n,n-2)]$ can be computed, and one continues in this way until eventually reaching $\ell=0$ for which $\kopt(n,0)=n$ holds.

Applying these computations to our example with $n=3$, we first easily obtain $\kopt(n=3,\ell=2)=1$ and $\E[T_{\opt}(n=3,\ell=2)]=3$, as in the drift maximizing case analyzed above. Given that $\kopt(3,0)=3$, the only interesting case is fitness level $\ell=1$. The expected remaining time $\E[T_1(n=3,\ell=1)]$ equals $1+\tfrac{2}{3}\E[T_{\opt}(3,2)]+\tfrac{1}{3}\E[T_1(3,1)]$. Since $\E[T_{\opt}(3,2)]=3$, a simple algebraic transformation shows $\E[T_{1}(3,1)]=\tfrac{9}{2}$. When flipping two bits, we either obtain the optimal solution (this happens with probability $1/3$) or we remain at the current fitness level, which shows that $\E[T_2(3,1)]=1+\tfrac{2}{3}\E[T_2(3,1)]$. Thus, $\E[T_2(3,1)]=3$. Finally, we compute that $\E[T_3(3,1)]=1+\E[T_{\opt}(3,2)]=4$. We therefore see that $\kopt(n=3,\ell=1)=2$ and $\E[T_{\opt}(3,1)]=3$. With these values, we obtain that the expected optimization time of \RLSopt on the 3-dimensional \onemax is $1+\tfrac{3}{8}\E[T_{\opt}(3,2)]+ \tfrac{3}{8}\E[T_{\opt}(3,1)]+\tfrac{1}{8}=\tfrac{27}{8}=3.375$.

\textbf{Optimal Mutation Strengths Need Not be Uneven.} With this example, we not only prove that $\fRLSopt \neq \fRLSdrift$, but we also make another interesting observation, which concerns the parity of the values $\kopt(n,\ell)$. It was proven in~\cite{DoerrDY20} that $\kdrift$ takes only odd values, since for every $k$ the drift of flipping $2k$ bits is strictly smaller than that of flipping $2k+1$ bits. 

The example above shows that the situation is different for $\kopt$. More precisely, we have seen that in the situation $n=3$ and $\ell=1$ flipping 2 bits is optimal.

\subsection{\texorpdfstring{$\foeaopt \neq \foeadrift$}{{(1+1)} EA-opt is not equal to {(1+1)} EA-drift} for \texorpdfstring{$n=3$}{n=3}}
\label{sec:oea3}

The only difference between the \oea and RLS is the random choice of the mutation strength~$k$, which the \oea samples from a binomial distribution $\Bin(n,p)$. 
The \oeadrift is defined by choosing the fitness-dependent mutation rate $\pdrift(n,\ell)$ which maximizes the expected progress 
\begin{align}\label{def:driftp}
& \E[\Delta(n,\ell,p)] \\
&\nonumber : =  \sum_{i=\ell+1}^{n}{(i-\ell) \Pr[\OM(y)=i \mid \OM(x)=\ell, y \assign \flip_k(x), k\sim \Bin(n,p)]}\\
&\nonumber \, = \sum_{k=1}^{n}{\Bin(n,p)(k) \E[\Delta(n,\ell,k)]}\\
&\nonumber \, = \sum_{k=1}^{n}{ \left(
	\binom{n}{k}p^{k}(1-p)^{n-k}
		\sum_{i=\lceil k/2 \rceil}^{k}
		\frac{\binom{n-\ell}{i}\binom{\ell}{k-i}\left(2i-k\right)}{\binom{n}{k}}
		\right)
		},
\end{align}
where we recall that $\E[\Delta(n,\ell,k)]$ had been defined in equation~\eqref{eq:Edriftk}.

Following the same arguments as in the definition of \RLSopt in Section~\ref{sec:RLS3}, the \oeaopt is defined by choosing in each fitness level the mutation rate $\popt(n,\ell)$ which minimizes the expected remaining time, i.e., the expression 
\begin{align*}
\E[T_{p}(n,\ell)]
= 
1 & + \Pr[\OM(y) \le \ell \mid \mathcal{E}] \E[T_{p}(n,\ell)]\\
&+\sum_{i=\ell+1}^{n-1}{\Pr[\OM(y)=i \mid \mathcal{E}] \E[T_{\opt}(n,i)]},
\end{align*}
where we abbreviate by $\mathcal{E}$ the event that $\OM(x)=\ell$, $y \assign \flip_{k}(x)$, and $k\sim \Bin(n,p)$. 

As above, we thus need to determine first the values of $\popt(n,n-1)$ and $\E[T_{\opt}(n,n-1)]$, then progress with the computation of $\popt(n,n-2)$ and $\E[T_{\opt}(n,n-2)]$, etc.

For $n=3$ we obtain the following values, which prove that, like for RLS, drift-maximization is also not optimal for the \oea.

\vspace{2ex}
\begin{tabular}{l|l|cc|cc}
$\ell$                   & $p^0(\ell)$ & $\pdrift(\ell)$ & $\E[T_{\drift}(\ell)]$ & $\popt(\ell)$ & $\E[T_{\opt}(\ell)]$ \\
\hline
3  & $1/8$ & -     & 0     & -    	& 0 \\
2  & $3/8$ & $1/3$ & 6.75  & $1/3$ & 3 \\
1  & $3/8$ & $1$   & 7.75  & $2/3$ & 3 \\
0  & $1/8$ & $1$   & 1     & $1$   & 1 \\
\hline 
 \multicolumn{2}{c|}{$\E[T]$} &    \multicolumn{2}{c|}{6.5625} &   \multicolumn{2}{c}{6.1875}                  
\end{tabular} 
\vspace{2ex}

Another interesting observation that we can make by comparing this table with the corresponding one of RLS (Sec.~\ref{sec:RLS3}) is that $\pdrift(3,\ell)=\kdrift(3,\ell)/3$ and $\popt(3,\ell)=\kopt(3,\ell)/3$. We will discuss this effect in more detail in Section~\ref{sec:comppkopt}.

\subsection{\texorpdfstring{$\foearesopt \neq \foearesdrift$}{{(1+1)} EA-opt is not equal to {(1+1)} EA-drift} for \texorpdfstring{$n=3$}{n=3}}
\label{sec:oeares3}

\oearesopt and \oearesdrift are defined by replacing in all definitions in Section~\ref{sec:oea3} the binomial distribution $\Bin(n,p)$ by the conditional binomial distribution $\Bin_{>0}(n,p)$, and by replacing the formulas accordingly. We omit a detailed definition for reasons of space. All replacements are straightforward, the only particularity to pay attention to is that both the  drift and the expected remaining time may be better for ever smaller values of $p$. This happens when flipping one bit  deterministically is better in terms of drift or expected running time, respectively, than using standard bit mutation. In this case we can either use the convention that the conditional standard bit mutation with mutation rate $p=0$ is to be interpreted as the $\flip_1$ operator (i.e., we set $\Bin(n,0)(1)=1$ and $\Bin(n,0)(k)=0$ for all $k \neq 1$), or we set a lower bound $\pmin$ for the mutation rate. The effects of the lower bound will be discussed in Section~\ref{sec:runtime}. 
When using $\pmin=0$, the situation for the \oeares for \onemax in dimension $n=3$ is given by the following table.

\vspace{2ex} \noindent
\begin{tabular}{l|l|cc|cc}
$\ell$                   & $p^0(\ell)$ & $p_{>0,\text{drift}}(\ell)$ & $\E[T_{>0,\text{drift}}(\ell)]$ & $p_{>0,\text{opt}}(\ell)$ & $\E[T_{>0,\text{opt}}(\ell)]$ \\
\hline
3  & $1/8$ & -     & 0     & -    	& 0 \\
2  & $3/8$ & 0 & 3 & 0 & 3 \\
1  & $3/8$ & $1$   & 4  & $3/4$ & $27/7$ \\
0  & $1/8$ & $1$   & 1     & $1$   & 1 \\
\hline 
 \multicolumn{2}{c|}{$\E[T]$} &    \multicolumn{2}{c|}{3.75} &   \multicolumn{2}{c}{$\approx 3.696$}                  
\end{tabular} 
\vspace{2ex}


\section{Optimal RLS and {(1+1)} EA Variants}
\label{sec:opt}

Using the formulas provided in Section~\ref{sec:example} we can compute the optimal RLS, \oea, and \oeares algorithms, as well as their drift-maximizing counterparts. Note, though, that the numerical evaluation of the binomial coefficients, as well as the optimization required to determine $\popt$ and $\presopt$ is not straightforward. For the latter, we have used the bounded method of the \emph{scipy} optimization module~\cite{scipy}. The overall expected running times are summarized in Table~\ref{tab:runtimes}, which can be found at the end of this paper.


\subsection{Optimal Mutation Strengths}
\label{sec:kopt}

\begin{figure}
\centering
\includegraphics[width=0.6\linewidth]{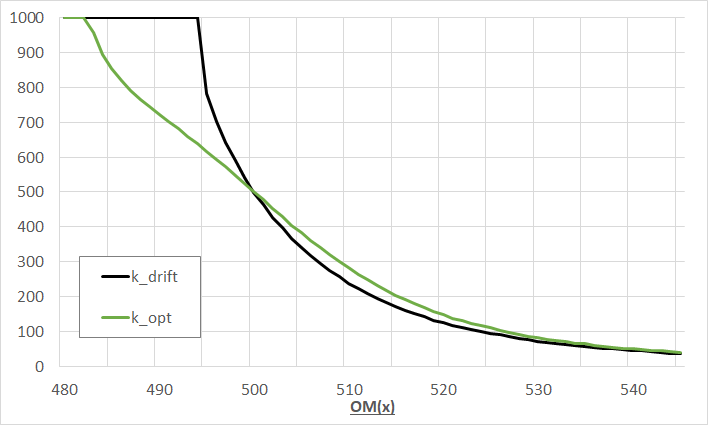}
\caption{Comparison of $\kopt(n,\ell)$ and $\kdrift(n,\ell)$ for $n=1{,}000$ (zoom into fitness levels $480 \le \ell \le 545$).}
\label{fig:kpoptdrift}
\end{figure}

We start our comparison by considering the differences between the drift-maximizing and the optimal mutation strengths for RLS. 
Figure~\ref{fig:kpoptdrift} plots the interesting region of $\kopt(n,\ell)$ and $\kdrift(n,\ell)$ for $n=1{,}000$; the overall picture is very similar across all dimensions $n$. In particular it holds for all $n$ that the curves cross at fitness level $\ell=n/2$. For smaller values, the optimal mutation strengths are smaller or identical to drift-maximizing ones, and the situation is reversed for fitness levels $\ell>n/2$. This can be explained by the formulas given in Section~\ref{sec:example}. While the drift-maximizer values a potential progress of $i$ by this same value, regardless of the current fitness level, the same potential progress is valued by $\E[T_{\opt}(\ell+i)]-\E[T_{\opt}(\ell)]>i$. \RLSdrift is thus more risk-averse than \RLSopt. Put differently, the latter makes use of the fact that an unlikely large fitness gain results in a larger reduction of the expected remaining optimization time than a more likely small fitness increase. \RLSopt therefore accepts a smaller probability of an improving move, at the benefit of a potentially larger fitness increase. This observation also explains why the extreme-valued parameter adaptation method proposed in~\cite{FialhoCSS08} showed better performance on \onemax than update schemes based on average gains. 

It was proven in~\cite{DoerrDY20} that an approximated drift-maximizer always flips only one bit when $\ell> 2n/3$. For the actual drift-maximizer this has not been formally proven, but in all our numerical evaluations for dimensions up to $10{,}000$ we have $\kopt(\ell)=\kdrift(\ell)=1$ for $\ell>2n/3$.

For dimension $n=1{,}000$, we see from Fig.~\ref{fig:kpoptdrift} that $\kopt(\ell)=1{,}000$ for $\ell\le 482$ and $\kdrift(\ell)=1{,}000$ for $\ell\le 494$. In this regime it is thus beneficial to invest one iteration to obtain, deterministically, a search point with function value $n-\ell$. 

The difference between the two functions becomes negligible for $\ell>545$. 

We do not plot the comparison of $\popt$ with $\pdrift$ nor that of $\presopt$ vs. $\presdrift$; their curves, however, are similar to those of RLS.  

\subsection{Comparison of \texorpdfstring{$\kopt$}{k-opt} and \texorpdfstring{$\popt$}{p-opt}}
\label{sec:comppkopt}

We have observed in Section~\ref{sec:oea3} that for $n=3$ the values of $\popt$ were identical to $\kopt/n$. Likewise, we had observed that in this example $\pdrift=\kdrift/n$. Figure~\ref{fig:kpopt10k} plots $\kopt(n,\ell)$ and $n\popt(n,\ell)$ for $n=10{,}000$ and Figure~\ref{fig:kpopt1k} plots $\kdrift(n,\ell)$ and $n\pdrift(n,\ell)$ for $n=1{,}000$; the overall picture is the same for drift-maximizing and optimal functions in both cases. 

While Figure~\ref{fig:kpopt10k} gives the global picture, Figures~\ref{fig:kpopt1k} zooms into the region in which $\kdrift$ is between 3 and 47. We observe that the mutation strength is always smaller, but very close to $n$ times the respective mutation rate. At the points at which $\kopt$ and $\kdrift$ change value the difference between $n\popt$ and $n\pdrift$ is smallest. 

\begin{figure}
\centering
\includegraphics[width=0.6\linewidth]{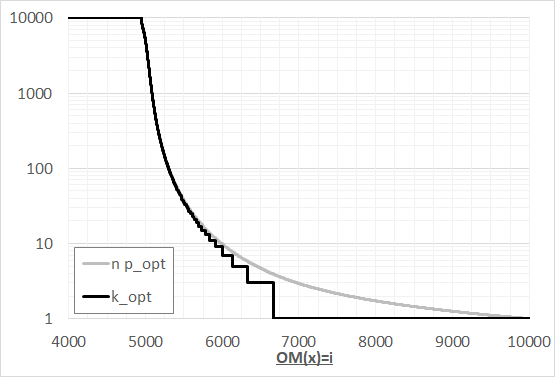}
\caption{Comparison of $\kopt(n,\ell)$ and $n\popt(n,\ell)$ for $n=10{,}000$. Note the logarithmic scale.}
\label{fig:kpopt10k}
\end{figure}

\begin{figure}
\centering
\includegraphics[width=0.6\linewidth]{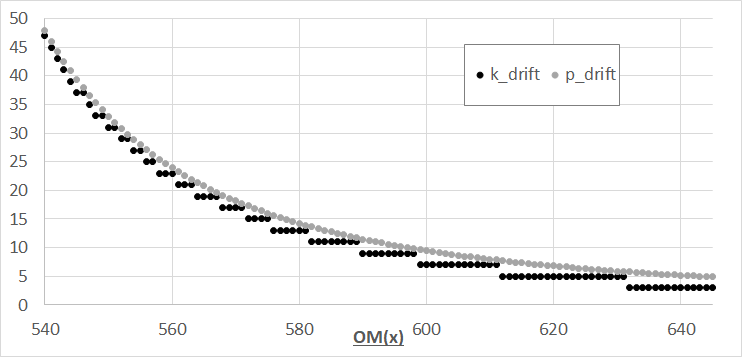}
\caption{Comparison of $\kdrift(n,\ell)$ and $n\pdrift(n=,\ell)$ for $n=1{,}000$ (zoom into fitness levels $540 \le \ell \le 645$).}
\label{fig:kpopt1k}
\end{figure}


\section{Running Times}
\label{sec:runtime}

We now discuss the impact of the differences in mutation strengths and rates on the overall expected running times.

\begin{figure}
\centering
\includegraphics[width=0.6\linewidth]{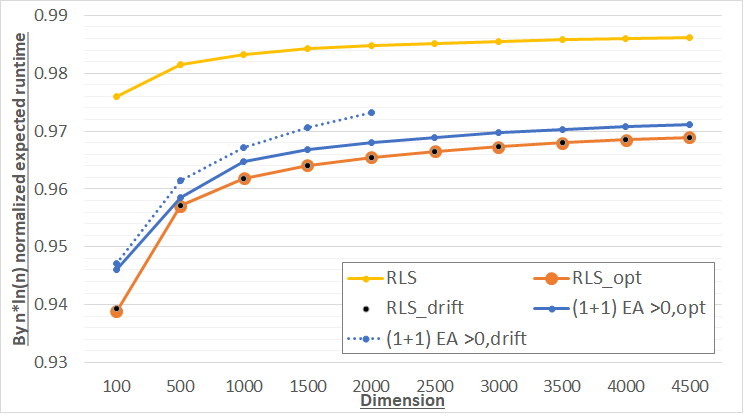}
\caption{Expected optimization time of different variants of RLS and the \oearesopt, normalized by $n \ln(n)$.}
\label{fig:runtimeRLS}
\end{figure}

We start our comparison with the RLS and the \oeares variants. Figure~\ref{fig:runtimeRLS} plots the by $n\ln(n)$ normalized optimization times of five different algorithms for 10 different problem dimensions between 100 and $4{,}500$. We denote here and in the following by RLS the traditional RLS variant using static mutation strength~$k=1$. We see that there is practically no difference between \RLSopt and \RLSdrift, and this despite the significant differences in the mutation strengths $\kopt$ and $\kdrift$. While the asymptotic result from~\cite{DoerrDY20} guarantees that the absolute difference is bounded by $O(n^{2/3} \log^9(n))$, the absolute difference between the two algorithms is even less than 1 across all tested problem dimensions. The normalized running times of both algorithms increase from around $0.939$ for $n=100$ to around $0.969$ for $n=4{,}500$. As we know from the theoretic result~\cite{DoerrDY20} these values converge to 1 for growing dimension~$n$.  

The differences between the \oearesopt and the \oearesdrift to \RLSopt are very small. The difference between the first two algorithms seems to be more significant than between drift-maximizing and optimal RLS variants, 
with a numerical difference between \oearesopt and \oearesdrift of around $0.5\%$ for $n=2{,}000$. We do not have an explanation for this comparatively large difference, but it may be caused by the numerical precision at which the results have been computed. More details about the \oeares will be discussed in Section~\ref{sec:influence}.

\begin{figure}
\centering
\includegraphics[width=0.6\linewidth]{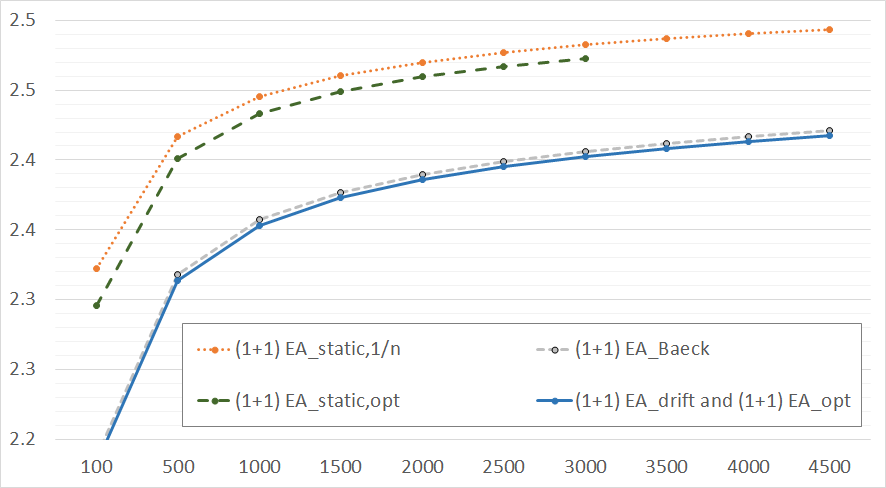}
\caption{Expected optimization time of different variants of \oea, normalized by $n \ln(n)$.}
\label{fig:runtimeoea}
\end{figure}

Our next chart, Figure~\ref{fig:runtimeoea}, compares the expected running times of different \oea variants. We first note that we plot two different static versions, one using the asymptotically optimal static mutation rate $1/n$, and the other one using the optimal static mutation rate per each dimension. The latter is slightly larger than $1/n$, as was already proven in~\cite{ChicanoWA14}. Since they only computed the optimal static rates for $n \le 100$, we also had to compute these for larger dimensions (using a direct computation, not the there-suggested matrix-based approach). Alternatively, we could have used the approximations suggested in~\cite{GiessenW18}, which extend the results of~\cite{ChicanoWA14} to the ${(1+\lambda)}$~EA and to larger dimensions. The relative advantage over $1/n$ is not very pronounced, and decreases from around $1.2\%$ for $n=100$ to around $0.4\%$ for $n=3{,}000$. The curves of the \oeadrift and the \oeares are  practically indistinguishable in this plot. Like for RLS the absolute difference between the expected running time of the two algorithms is less than 1 for all tested dimensions, again despite significant differences in the functions $\popt$ and $\pdrift$. We add to this chart a comparison with the \oea using the fitness-dependent mutation rate $p(\ell)=1/(2\ell+2-n)$ (for $\ell \ge n/2$) suggested in~\cite{Back93}; we use $p(\ell)=\pdrift(\ell)$ for $\ell<n/2$. B\"ack obtained this mutation rate from numerical evaluations of $\pdrift$ in small dimensions $n\le 100$. His algorithm performs only slightly worse than the true drift-maximizing \oeadrift, and, thus, as the \oeaopt.

\subsection{Influence of \texorpdfstring{$\pmin$}{pmin} on the \texorpdfstring{\oeares}{Resampling {(1+1)} EA}}
\label{sec:influence} 

We have briefly mentioned in Section~\ref{sec:oeares3} that for the \oeares one needs to specify a lower bound for the mutation probability, since in some situations the optimal mutation rate is zero (when using the convention that $\Bin(n,0)$ deterministically returns one). For practical applications such small mutation rates may be undesirable, e.g., when using multiplicative success-based updates rules as suggested in~\cite{DoerrW18}. We therefore investigate the influence of this lower bound on the expected running times. These normalized running times are plotted for six different algorithms in Figure~\ref{fig:runtimeoeares}. The drift-maximizing variants would be indistinguishable in this plot from the optimal ones, and are therefore omitted, except for the case $\pmin=0$, which we have already discussed in Figure~\ref{fig:runtimeRLS}. Note that the \oeares with optimal static mutation rate uses $\pmin=0$, and is therefore equal to RLS. The relative disadvantage of increasing $\pmin$ to $1/(2n)$ increases from around $22\%$ in dimension $n=100$ to around $26\%$ in dimension 3{,}500, both for the static and the adaptive variants. Further increasing $\pmin$ to $1/n$ results in a relative disadvantage of $51-59\%$ for the static and from $52-60\%$ for the dynamic variants.

\begin{figure}
\centering
\includegraphics[width=0.6\linewidth]{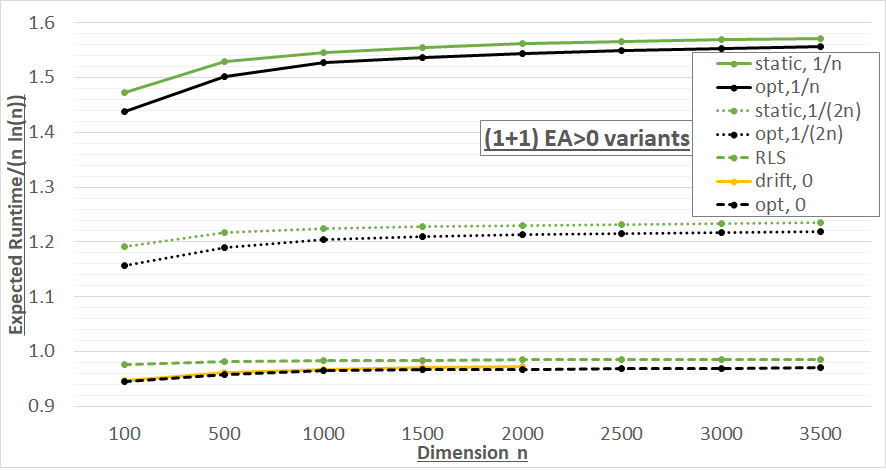}
\caption{Expected Optimization time of different variants of \oeares, normalized by $n \ln(n)$.}
\label{fig:runtimeoeares}
\end{figure}

\subsection{Anytime Performance}
\label{sec:anytime}

\textbf{Fixed-Budget Results.} 
While we have focused above on expected optimization times 
we will now follow the suggestion made in~\cite{DoerrDY20} and provide a more detailed analysis of the \emph{anytime behavior} of the algorithms. More precisely, we regard fixed-budget performance of \RLSopt, \RLSdrift, and RLS. Only \RLSopt and RLS are plotted in Figure~\ref{fig:FB}, the curves of \RLSopt and \RLSdrift are practically indistinguishable. Note that the numbers underlying the plot in Figures~\ref{fig:FB} and~\ref{fig:FT} (discussed in the next section) are the only ones in this paper that are not derived from theoretical bounds. We have performed a simulation of 500 independent runs of the three algorithms instead, and we used IOHprofiler~\cite{IOHprofiler} to analyze the runtime data. We show not only the mean value, but also the standard deviation. The curves are well separated even when considering these, for all budgets up to around $3{,}500$. Analyzing the data in more detail, we observe that the relative advantage in average function value decreases from 10\% for budget 100 to 1\% for budget $2{,}500$. For larger budgets, the average fitness value is less than 1\% larger for \RLSopt than for RLS. However, as proven to hold in an asymptotic sense for the \RLSdrift in~\cite{DoerrDY20}, the average distance to the optimum is constantly about $12-14\%$ better for \RLSopt than for RLS, for budgets up to $3{,}500$. The average function values at this budget ($3{,}500$ function evaluations) are slightly smaller than 990 for all three algorithms, RLS, \RLSopt, and \RLSdrift. For larger budgets, the distance to the optimum is hence very small. This, in combination with the variance of our simulation, results in inconsistent relative advantages in terms of distance to the optimum for budgets greater than $3{,}500$. 

\begin{figure}
\centering
\includegraphics[width=0.6\linewidth]{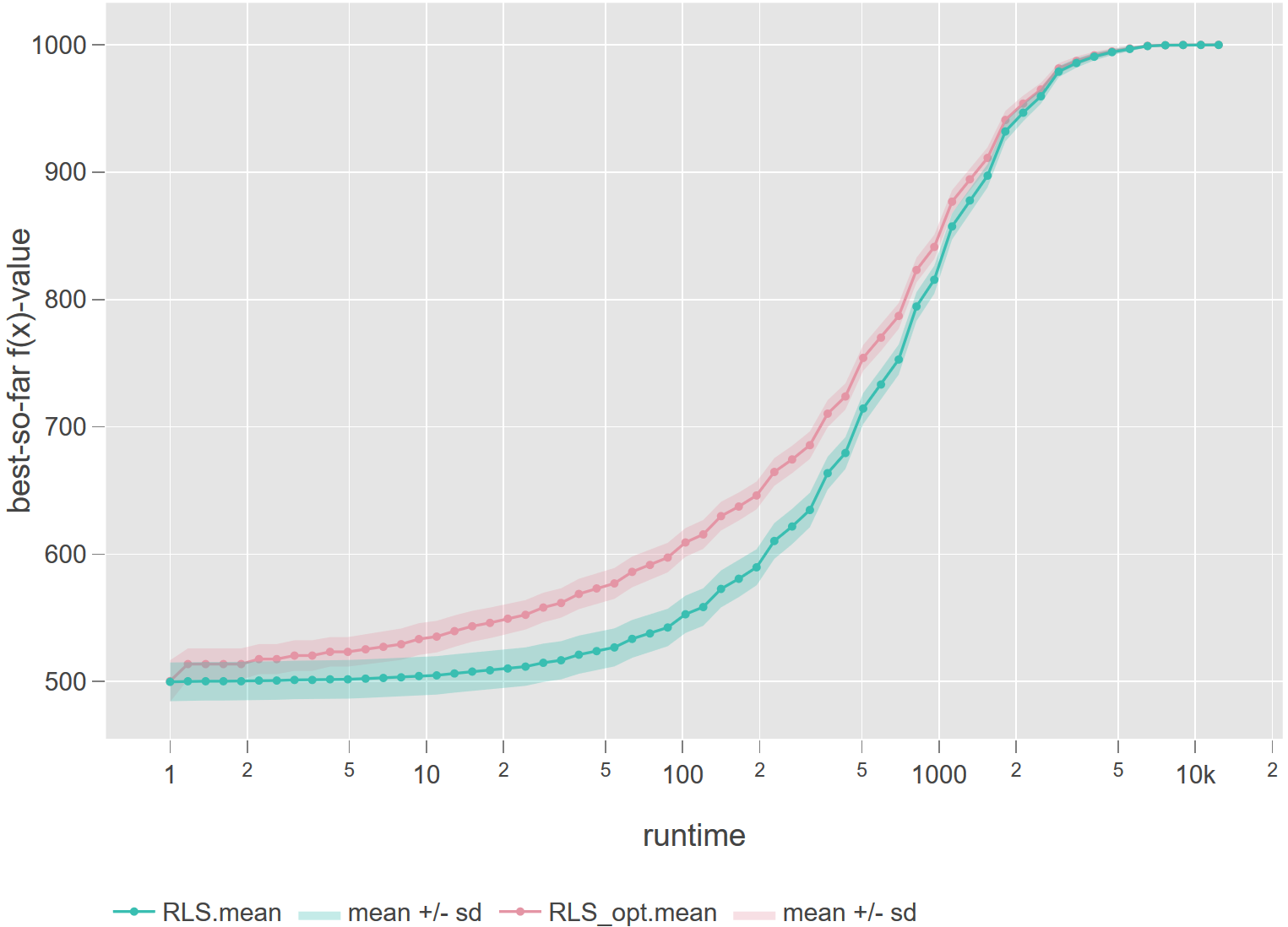}
\caption{Average fixed-budget results for RLS and \RLSopt on \onemax in dimension $n=1{,}000$ across 500 independent runs.}
\label{fig:FB}
\end{figure}

\textbf{Fixed-Target Results.} 
Using the same runtime data for the 500 runs, we can also compute fixed-target results, i.e., the function mapping each fitness level $\ell$ to the expected time needed to reach a solution $x$ of fitness $\OM(x)\ge \ell$. These values, of course, could also easily be computed theoretically from the results presented in Section~\ref{sec:kopt}, but we feel that the precision of the simulation suffices to demonstrate the main effects. The results are plotted in Figure~\ref{fig:FT}.

\begin{figure}
\centering
\includegraphics[width=0.6\linewidth]{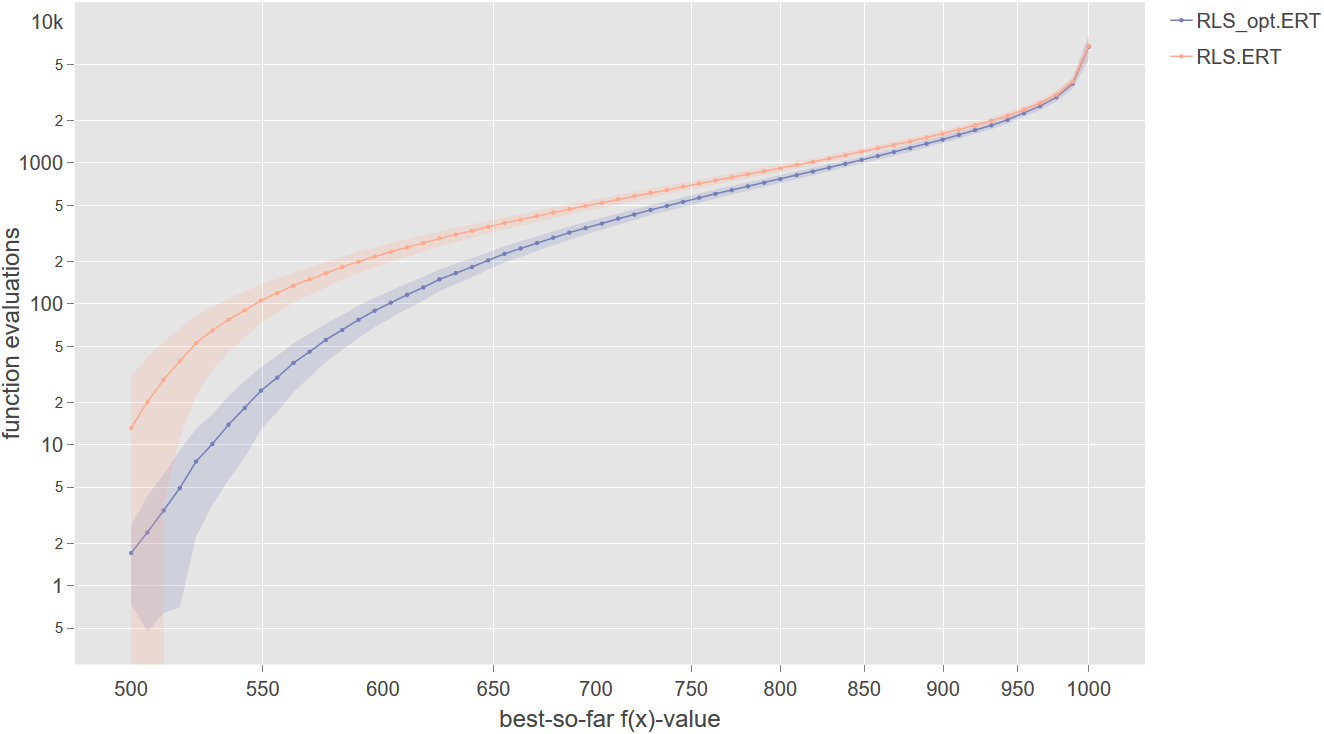}
\caption{Average fixed-target running time for RLS and \RLSopt on \onemax in dimension $n=1{,}000$ across 500 independent runs.}
\label{fig:FT}
\end{figure}

It is not difficult to see that \RLSopt is not optimal for minimizing the expected first hitting time of targets $\ell<n$, simply because overshooting the target $\ell$ are disadvantageous for this optimization goal. For a similar reason, \RLSopt is also not optimal in terms of maximizing the expected function value at a given budget of $B<\E[T(\fRLSopt)]$, i.e., when the budget is less than the expected overall optimization time of \RLSopt.

\subsection{Remaining Optimization Times}
\label{sec:remaining}

Finally, we take a look at the evolution of the expected remaining optimization time per each fitness level. These values, derived from our numerical evaluation of the theoretical bounds presented in Section~\ref{sec:example}, are plotted in Figure~\ref{fig:remaining}. While the algorithms with static mutation rates and strength are not able to profit from the fact that $\OM(\bar{x})=n-\OM(x)$ for each $x \in \{0,1\}^n$, we see an almost symmetric behavior for the adaptive algorithms. We also see again the influence of the lower bound $\pmin \in \{1/n, 1/(2n)\}$ in the \oeares variants, which are quite significant. 

\begin{figure}
\centering
\includegraphics[width=0.75\linewidth]{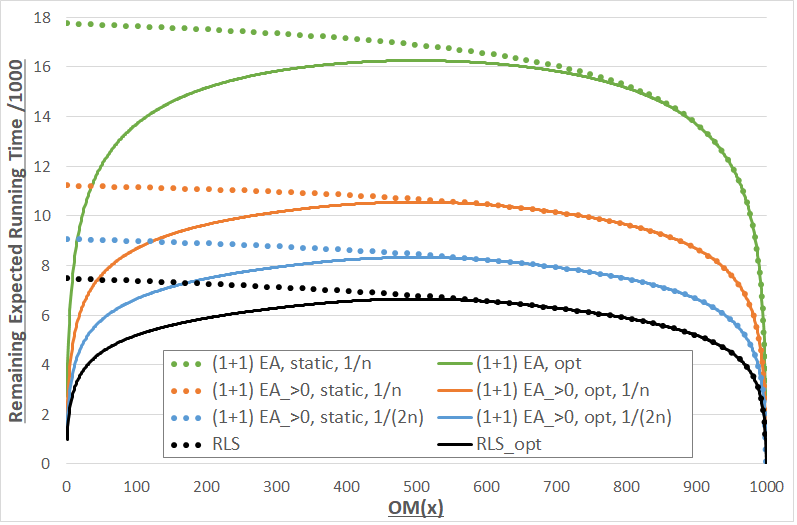}
\caption{Expected remaining optimization times for \onemax in dimension $n=1{,}000$.}
\label{fig:remaining}
\end{figure}

From this figure we can also compute the weights $\E[T_{\opt}(\ell+1)]-\E[T_{\opt}(\ell)]$ by which the \RLSopt starting in a search point of fitness $\ell$ values a potential fitness progress of $i$. We plot in Figure~\ref{fig:gradient} the gradient of 
the curves $\E[T(\ell)]$ plotted in Figure~\ref{fig:remaining}. 
That is, for every $\ell$ we plot the values $\E[T(1000,\ell)]-\E[T(1000,\ell-1)]$ for \RLSopt and RLS. We recall that \RLSdrift values a potential fitness progress of $i$ by the same value $i$. 
We thus clearly see that \RLSopt gives much more importance to large fitness gains, and hence uses the already discussed more risky strategy aiming at potentially larger fitness gains, at the cost of a larger probability of creating an offspring that will be discarded.

\subsection{Best Unary Unbiased Algorithms for OneMax}

Note that plot in Figure~\ref{fig:remaining} also raises the question how much the algorithms lose in performance by being forced to be elitist. Note that slightly better algorithms are possible when allowing them to first decrease the function value to 0 and then inverting the bit string. For the adaptive algorithms, this would clearly bring more flexibility, and a provable positive advantage over the elitist algorithms studied in this work. Put differently, the best unary unbiased black-box algorithm for \onemax is slightly better than \RLSopt. The almost perfect symmetric shape of the algorithms in Figure~\ref{fig:remaining}, however, indicates that the advantage is very small. A rigorous quantification, which we consider to be of rather philosophical benefit, is left for future work. 

\begin{figure}
\centering
\includegraphics[width=0.6\linewidth]{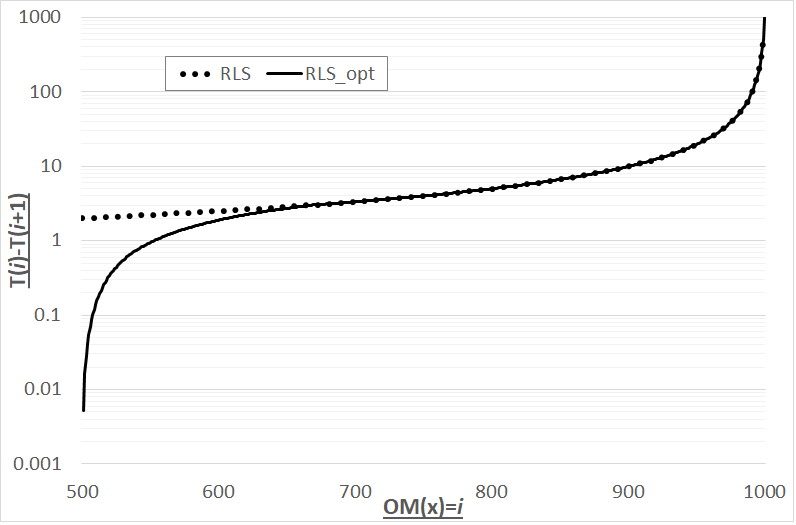}
\caption{Gradient of expected remaining optimization times for \onemax in dimension $n=1{,}000$.}
\label{fig:gradient}
\end{figure}

\section{Discussion}
\label{sec:conclusions}

We have shown that the assumption that drift-maximization is optimal for solving the \onemax problem is not correct, neither for RLS, nor the \oea, nor the \oeares. A more risky strategy turns out to be optimal. However, while the differences in the drift-maximizing and the optimal mutation rates are significant (Figure~\ref{fig:kpoptdrift}), the difference in expected running time is negligibly small already for very small dimensions. The structural findings made here for the \onemax problem also applies in a broader sense to the optimization of non-deceptive problems. Already for linear functions like {$\textsc{BinVal}$}, the difference between drift-maximizing and optimal RLS and \oea variants may be more substantial than for \onemax. We also note that, while we have restricted ourselves to {(1+1)}-type algorithms, similar effects also hold for population-based EAs. 

The computation of the drift-maximizing and time-minimizing mutation strengths and rates 
are quite tedious and require several days of computing time already for moderate dimension. 
In order to obtain valid baseline algorithms for larger dimensions, it would be desirable to derive closed formula expressions that approximate these functions sufficiently well. 
Note that the formula provided by B{\"a}ck for the drift-maximizer (cf. discussion in Section~\ref{sec:runtime}) 
seems to allow to derive quite reliable predictions for the drift-maximizing {(1+1)}~EA as seen in Figure~\ref{fig:runtimeoea}. 


\subsubsection*{Acknowledgments.}
We thank Thomas B\"ack for several valuable discussions on the history of adaptive parameter settings. 

Our research benefited from the support of 
the \emph{Paris Ile-de-France Region}, 
a public grant as part of the Investissement d'avenir project, reference \emph{ANR-11-LABX-0056-LMH}, LabEx LMH, in a joint call with Gaspard Monge Program for optimization, operations research and their interactions with data sciences, and 
\emph{COST Action CA15140} on 'Improving Applicability of Nature-Inspired Optimisation by Joining Theory and Practice (ImAppNIO)' supported by COST (European Cooperation in Science and Technology).

}

\begin{sidewaystable}[t]
\begin{tabular}{r|rrrrrrrrrr}
& \multicolumn{10}{c}{problem dimension} \\
\cmidrule(lr){2-11}
Algorithm & 100                                    & 500   & 1{,}000  & 1{,}500   & 2{,}000   & 2{,}500   & 3{,}000   & 3{,}500   & 4{,}000   & 4{,}500    \\
\hline
\RLSopt                               	& 	 433.4 	& 	 2,975.0 	& 	 6,645.0 	& 	 10,576.7 	& 	 14,678.4 	& 	 18,906.4 	& 	 23,235.2 	& 	 27,647.7 	& 	 32,131.9 	& 	 36,678.8 	\\
\RLSdrift 	& 	433.6 	& 	2,975.3 	 	& 	  6,645.2 	  	& 	  10,576.9 	  	& 	  14,678.6 	  	& 	  18,906.7 	  	& 	  23,235.5 	  	& 	  27,648.0 	  	& 	  32,132.2 	  	& 	 36,679.0 	\\
\cdashline{2-11}																					
RLS                                    	& 	 450 	& 	 3,051 	& 	 6,793 	& 	 10,797 	& 	 14,971 	& 	 19,272 	& 	 23,673 	& 	 28,158 	& 	 32,714 	& 	 37,333 	\\
\hline																					
{(1+1)} EA$_{>0,\opt}$           	& 	 437 	& 	 2,979 	& 	 6,665 	& 	 10,606 	& 	 14,717 	& 	 18,954 	& 	 23,292 	& 	 27,714 	& 	 32,207 	& 	 36,763 	\\
{(1+1)} EA$_{>0,\opt,\pmin=1/(2n)}$ 	& 	 534 	& 	 3,700 	& 	 8,321 	& 	 13,270 	& 	 18,441 	& 	 23,775 	& 	 29,239 	& 	 34,813 	& 		& 		\\
{(1+1)} EA$_{>0,\opt,\pmin=1/n}$ 	& 	 663 	& 	 4,666 	& 	 10,548 	& 	 16,865 	& 	 23,473 	& 	 30,298 	& 	 37,296 	& 	 44,438 	& 		& 		\\
\cdashline{2-11}																					
{(1+1)} EA$_{>0,\drift}$ 	& 	 437 	& 	 2,989 	& 	 6,682 	& 	 10,648 	& 	 14,795 	& 		& 		& 		& 		& 		\\
{(1+1)} EA$_{>0,\drift,\pmin=1/(2n)}$ 	& 	 534 	& 	 3,711 	& 	 8,322 	& 	 13,271 	& 	 18,442 	& 	 23,776 	& 	 29,242 	& 	 34,814 	& 		& 		\\
{(1+1)} EA$_{>0,\drift,\pmin=1/n}$ 	& 	 664 	& 	 4,682 	& 	 10,549 	& 	 16,865 	& 	 23,473 	& 	 30,298 	& 	 37,297 	& 	 44,438 	& 		& 		\\
\cdashline{2-11}																					
{(1+1)} EA$_{>0,\text{static},p=1/(2n)}$ 	& 	 550 	& 	 3,781 	& 	 8,458 	& 	 13,475 	& 	 18,712 	& 	 24,113 	& 	 29,644 	& 	 35,284 	& 		& 		\\
{(1+1)} EA$_{>0,\text{static},p=1/n}$ 	& 	 679 	& 	 4,751 	& 	 10,684 	& 	 17,066 	& 	 23,740 	& 	 30,631 	& 	 37,695 	& 	 44,902 	& 	 52,233 	& 	 59,672 	\\
\hline																					
\oeaopt                          	& 	 1,006 	& 	 7,189 	& 	 16,254 	& 	 26,031 	& 	 36,269 	& 	 46,850 	& 	 57,705 	& 	 68,788 	& 	 80,065 	& 		\\
\oeadrift                        	& 	 1,006 	& 	 7,189 	& 	 16,254 	& 	 26,031 	& 	 36,269 	& 	 46,850 	& 	 57,706 	& 	 68,788 	& 	 80,066 	& 	 91,513 	\\
{(1+1)} EA$_{\text{B\"ack}}$       	& 	 1,008 	& 	 7,203 	& 	 16,283 	& 	 26,075 	& 	 36,328 	& 	 46,925 	& 	 57,795 	& 	 68,893 	& 	 80,186 	& 	 91,649 	\\
\cdashline{2-11}																					
{(1+1)} EA$_{\text{static},p=\opt}$                  	& 	 1,058 	& 	 7,461 	& 	 16,807 	& 	 26,867 	& 	 37,389 	& 	 48,257 	& 	 59,398 	& 		& 		& 		\\
{(1+1)} EA$_{\text{static},p=1/n}$ 	& 	 1,071 	& 	 7,510 	& 	 16,896 	& 	 26,992 	& 	 37,550 	& 	 48,451 	& 	 59,626 	& 	 71,028 	& 	 82,625 	& 	 94,392 	\\

\end{tabular}
\caption{Expected Optimization Times of Different Variants of RLS and the {(1+1)}~EA on \onemax for problem dimensions between $n=100$ and $n=4{,}500$.}
\label{tab:runtimes}
\end{sidewaystable}

\small
%

\begin{thebibliography}{dPdLDD15}

\bibitem[AM16]{AletiM16}
Aldeida Aleti and Irene Moser.
\newblock A systematic literature review of adaptive parameter control methods
  for evolutionary algorithms.
\newblock {\em ACM Computing Surveys}, 49:56:1--56:35, 2016.

\bibitem[B{\"{a}}c92]{Back92}
Thomas B{\"{a}}ck.
\newblock The interaction of mutation rate, selection, and self-adaptation
  within a genetic algorithm.
\newblock In {\em Proc. of Parallel Problem Solving from Nature (PPSN'92)},
  pages 87--96. Elsevier, 1992.

\bibitem[B{\"a}c93]{Back93}
Thomas B{\"a}ck.
\newblock Optimal mutation rates in genetic search.
\newblock In {\em Proc. of the 5th International Conference on Genetic
  Algorithms (ICGA'93)}, pages 2--8. Morgan Kaufmann, 1993.

\bibitem[BD20]{BuzdalovD20}
Maxim Buzdalov and Carola Doerr.
\newblock Optimal mutation rates for the $(1+\lambda)$ {EA} on {O}ne{M}ax.
\newblock In {\em Proc. of Parallel Problem Solving from Nature (PPSN'20)},
  volume 12270 of {\em LNCS}, pages 574--587. Springer, 2020.

\bibitem[BLS14]{BadkobehLS14}
Golnaz Badkobeh, Per~Kristian Lehre, and Dirk Sudholt.
\newblock Unbiased black-box complexity of parallel search.
\newblock In {\em Proc. of Parallel Problem Solving from Nature (PPSN'14)},
  volume 8672 of {\em Lecture Notes in Computer Science}, pages 892--901.
  Springer, 2014.

\bibitem[CD18a]{CarvalhoD18}
Eduardo {Carvalho Pinto} and Carola Doerr.
\newblock A simple proof for the usefulness of crossover in black-box
  optimization.
\newblock In {\em Proc. of Parallel Problem Solving from Nature (PPSN'18)},
  volume 11102 of {\em Lecture Notes in Computer Science}, pages 29--41.
  Springer, 2018.
\newblock Full version available at \url{http://arxiv.org/abs/1812.00493}.

\bibitem[CD18b]{CarvalhoD17}
Eduardo {Carvalho Pinto} and Carola Doerr.
\newblock Towards a more practice-aware runtime analysis of evolutionary
  algorithms.
\newblock {\em CoRR}, abs/1812.00493, 2018.

\bibitem[CHJ{\etalchar{+}}17]{CorusHJOSZ17}
Dogan Corus, Jun He, Thomas Jansen, Pietro~Simone Oliveto, Dirk Sudholt, and
  Christine Zarges.
\newblock On easiest functions for mutation operators in bio-inspired
  optimisation.
\newblock {\em Algorithmica}, 78:714--740, 2017.

\bibitem[CWA14]{ChicanoWA14}
Francisco Chicano, Darrell Whitley, and Enrique Alba.
\newblock Exact computation of the expectation surfaces for uniform crossover
  along with bit-flip mutation.
\newblock {\em Theoretical Computer Science}, 545:76--93, 2014.

\bibitem[DD16]{DoerrD16impact}
Benjamin Doerr and Carola Doerr.
\newblock The impact of random initialization on the runtime of randomized
  search heuristics.
\newblock {\em Algorithmica}, 75:529--553, 2016.

\bibitem[DD18]{DoerrD18ga}
Benjamin Doerr and Carola Doerr.
\newblock Optimal static and self-adjusting parameter choices for the
  $(1+(\lambda,\lambda))$ genetic algorithm.
\newblock {\em Algorithmica}, 80:1658--1709, 2018.

\bibitem[DD20]{DoerrD18chapter}
Benjamin Doerr and Carola Doerr.
\newblock Theory of parameter control mechanisms for discrete black-box
  optimization: Provable performance gains through dynamic parameter choices.
\newblock In {\em Theory of Evolutionary Computation: Recent Developments in
  Discrete Optimization}, pages 271--321. Springer, 2020.

\bibitem[DDK18]{DoerrDK18}
Benjamin Doerr, Carola Doerr, and Timo K{\"{o}}tzing.
\newblock Static and self-adjusting mutation strengths for multi-valued
  decision variables.
\newblock {\em Algorithmica}, 80:1732--1768, 2018.

\bibitem[DDY16]{DoerrDY16PPSN}
Benjamin Doerr, Carola Doerr, and Jing Yang.
\newblock $k$-bit mutation with self-adjusting $k$ outperforms standard bit
  mutation.
\newblock In {\em Proc. of Parallel Problem Solving from Nature (PPSN'16)},
  volume 9921 of {\em Lecture Notes in Computer Science}, pages 824--834.
  Springer, 2016.

\bibitem[DDY20]{DoerrDY20}
Benjamin Doerr, Carola Doerr, and Jing Yang.
\newblock Optimal parameter choices via precise black-box analysis.
\newblock {\em Theoretical Computer Science}, 801:1--34, 2020.

\bibitem[DGWY19]{DoerrGWY19}
Benjamin Doerr, Christian Gie{\ss}en, Carsten Witt, and Jing Yang.
\newblock The $(1+\lambda)$ evolutionary algorithm with self-adjusting mutation
  rate.
\newblock {\em Algorithmica}, 81(2):593--631, 2019.

\bibitem[DJW12]{DoerrJW12}
Benjamin Doerr, Daniel Johannsen, and Carola Winzen.
\newblock Multiplicative drift analysis.
\newblock {\em Algorithmica}, 64:673--697, 2012.

\bibitem[DL17]{DoerrL17ECJ}
Carola Doerr and Johannes Lengler.
\newblock Introducing elitist black-box models: When does elitist behavior
  weaken the performance of evolutionary algorithms?
\newblock {\em Evolutionary Computation}, 25:587--606, 2017.

\bibitem[DLMN17]{FastGA17}
Benjamin Doerr, Huu~Phuoc Le, R{\'{e}}gis Makhmara, and Ta~Duy Nguyen.
\newblock Fast genetic algorithms.
\newblock In {\em Proc. of Genetic and Evolutionary Computation Conference
  (GECCO'17)}, pages 777--784. ACM, 2017.

\bibitem[DLOW18]{DoerrLOW18}
Benjamin Doerr, Andrei Lissovoi, Pietro~Simone Oliveto, and John~Alasdair
  Warwicker.
\newblock On the runtime analysis of selection hyper-heuristics with adaptive
  learning periods.
\newblock In {\em Proc. of Genetic and Evolutionary Computation Conference
  (GECCO'18)}, pages 1015--1022. ACM, 2018.

\bibitem[dPdLDD15]{LaillevaultDD15}
Axel de~Perthuis~de Laillevault, Benjamin Doerr, and Carola Doerr.
\newblock Money for nothing: Speeding up evolutionary algorithms through better
  initialization.
\newblock In {\em Proc. of Genetic and Evolutionary Computation Conference
  (GECCO'15)}, pages 815--822. ACM, 2015.

\bibitem[DW18]{DoerrW18}
Carola Doerr and Markus Wagner.
\newblock On the effectiveness of simple success-based parameter selection
  mechanisms for two classical discrete black-box optimization benchmark
  problems.
\newblock In {\em Proc. of Genetic and Evolutionary Computation Conference
  (GECCO'18)}, pages 943--950. {ACM}, 2018.

\bibitem[DWY18a]{DoerrWY18}
Benjamin Doerr, Carsten Witt, and Jing Yang.
\newblock Runtime analysis for self-adaptive mutation rates.
\newblock In {\em Proc. of Genetic and Evolutionary Computation Conference
  (GECCO'18)}, pages 1475--1482. ACM, 2018.

\bibitem[DWY{\etalchar{+}}18b]{IOHprofiler}
Carola Doerr, Hao Wang, Furong Ye, Sander van Rijn, and Thomas B{\"a}ck.
\newblock {IOHprofiler: A Benchmarking and Profiling Tool for Iterative
  Optimization Heuristics}.
\newblock {\em arXiv e-prints:1810.05281}, October 2018.
\newblock IOHprofiler is available at \url{https://github.com/IOHprofiler}.

\bibitem[FCSS08]{FialhoCSS08}
{\'{A}}lvaro Fialho, Lu{\'{\i}}s~Da Costa, Marc Schoenauer, and Mich{\`{e}}le
  Sebag.
\newblock Extreme value based adaptive operator selection.
\newblock In {\em Proc. of Parallel Problem Solving from Nature (PPSN'08)},
  volume 5199 of {\em Lecture Notes in Computer Science}, pages 175--184.
  Springer, 2008.

\bibitem[FCSS09]{FialhoCSS09}
{\'{A}}lvaro Fialho, Lu{\'{\i}}s~Da Costa, Marc Schoenauer, and Mich{\`{e}}le
  Sebag.
\newblock Dynamic multi-armed bandits and extreme value-based rewards for
  adaptive operator selection in evolutionary algorithms.
\newblock In {\em Proc. of Learning and Intelligent Optimization (LION'09)},
  volume 5851 of {\em Lecture Notes in Computer Science}, pages 176--190.
  Springer, 2009.

\bibitem[GW18]{GiessenW18}
Christian Gie{\ss}en and Carsten Witt.
\newblock Optimal mutation rates for the (1+$\lambda$) {EA} on {O}ne{M}ax
  through asymptotically tight drift analysis.
\newblock {\em Algorithmica}, 80(5):1710--1731, 2018.

\bibitem[HPR{\etalchar{+}}18]{HwangPRTC18}
Hsien{-}Kuei Hwang, Alois Panholzer, Nicolas Rolin, Tsung{-}Hsi Tsai, and
  Wei{-}Mei Chen.
\newblock Probabilistic analysis of the (1+1)-evolutionary algorithm.
\newblock {\em Evolutionary Computation}, 26, 2018.

\bibitem[HW19]{HwangW19}
Hsien{-}Kuei Hwang and Carsten Witt.
\newblock Sharp bounds on the runtime of the {(1+1)} {EA} via drift analysis
  and analytic combinatorial tools.
\newblock In {\em Proc. of {ACM/SIGEVO} Conference on Foundations of Genetic
  Algorithms (FOGA'19)}, pages 1--12. ACM, 2019.

\bibitem[JOP{\etalchar{+}}  ]{scipy}
Eric Jones, Travis Oliphant, Pearu Peterson, et~al.
\newblock {SciPy}: Open source scientific tools for {Python}, 2001--.

\bibitem[KHE15]{KarafotiasHE15}
Giorgos Karafotias, Mark Hoogendoorn, and A.E. Eiben.
\newblock Parameter control in evolutionary algorithms: Trends and challenges.
\newblock {\em IEEE Transactions on Evolutionary Computation}, 19:167--187,
  2015.

\bibitem[LOW20]{LissovoiOW20}
Andrei Lissovoi, Pietro~S. Oliveto, and John~Alasdair Warwicker.
\newblock Simple hyper-heuristics control the neighbourhood size of randomised
  local search optimally for {LeadingOnes}.
\newblock {\em Evolutionary Computation}, 28(3):437--461, 2020.

\bibitem[LS11]{LassigS11}
J{\"o}rg L{\"a}ssig and Dirk Sudholt.
\newblock Adaptive population models for offspring populations and parallel
  evolutionary algorithms.
\newblock In {\em Proc. of Foundations of Genetic Algorithms (FOGA'11)}, pages
  181--192. ACM, 2011.

\bibitem[Sud13]{Sudholt13}
Dirk Sudholt.
\newblock A new method for lower bounds on the running time of evolutionary
  algorithms.
\newblock {\em IEEE Transactions on Evolutionary Computation}, 17:418--435,
  2013.

\bibitem[Thi09]{Thierens09}
Dirk Thierens.
\newblock On benchmark properties for adaptive operator selection.
\newblock In {\em Proc. of Genetic and Evolutionary Computation Conference
  (GECCO'09), Companion Material}, pages 2217--2218. ACM, 2009.

\bibitem[Wit13]{Witt13j}
Carsten Witt.
\newblock Tight bounds on the optimization time of a randomized search
  heuristic on linear functions.
\newblock {\em Combinatorics, Probability {\&} Computing}, 22:294--318, 2013.

\bibitem[YDB19]{YeDB19}
Furong Ye, Carola Doerr, and Thomas B{\"{a}}ck.
\newblock Interpolating local and global search by controlling the variance of
  standard bit mutation.
\newblock In {\em Proc. of {IEEE} Congress on Evolutionary Computation
  (CEC'19)}, pages 2292--2299. IEEE, 2019.

\end{thebibliography}
\newcommand{\etalchar}[1]{$^{#1}$}

\end{document}